%% file: 0-main.tex
\documentclass[conference]{IEEEtran}
\usepackage{times}

\usepackage[numbers]{natbib}
\usepackage{multicol}
\usepackage[bookmarks=true]{hyperref}
\usepackage{multirow}

\usepackage{hyperref}   
\usepackage{url}

\usepackage{mathtools}
\usepackage{amsfonts}  % for \mathbb    
\usepackage{graphics}
\usepackage[pdftex]{graphicx}
\usepackage{wrapfig}
\usepackage{caption}
\usepackage{color}
\usepackage{dsfont}
\usepackage[]{mdframed}
\usepackage{algorithm}
\usepackage{algorithmic}
\usepackage{xparse}
\usepackage{amsmath}
\usepackage{mathtools}
\usepackage{amssymb}
\usepackage{amsthm}
\usepackage{amssymb}

\usepackage{booktabs}
\usepackage{xspace}
\usepackage{cleveref}
\usepackage{lipsum}

\usepackage[table]{xcolor}   % add in preamble
\usepackage{booktabs}        % cleaner rules
\usepackage{makecell}

\usepackage{xcolor} % For defining colors
\usepackage{hyperref} % For hyperlinks
\hypersetup{
    colorlinks=true,
}

\DeclarePairedDelimiterX{\infdivx}[2]{(}{)}{%
  #1\;\delimsize\|\;#2%
}

\usepackage{expl3}
\ExplSyntaxOn
\newcommand\latinabbrev[1]{
  \peek_meaning:NTF . {% Same as \@ifnextchar
    #1\@}%
  { \peek_catcode:NTF a {% Check whether next char has same catcode as \'a, i.e., is a letter
      #1.\@ }%
    {#1.\@}}}
\ExplSyntaxOff
\def\eg{\latinabbrev{e.g}}

\def\etc{\latinabbrev{etc}}

\NewDocumentCommand \proposition {g g g g} {\texttt{#1}(#2
  \IfValueTF{#3}{,\,#3}{}
  \IfValueTF{#4}{,\,#4}{}
  )
}

\NewDocumentCommand \actioncall {g g g g} {\text{#1}(#2
  \IfValueTF{#3}{,#3}{}
  \IfValueTF{#4}{,#4}{}
  \texttt{)}
}

\crefname{section}{Sec.}{Secs.}
\crefname{figure}{Fig.}{Figs.}
\crefname{table}{Tab.}{Tabs.}
\crefname{equation}{Eq.}{Eqs.}

\def \MethodName {Structured Affordance Grounding for Action\xspace}
\def \MethodAcronym {SAGA\xspace}

\IEEEoverridecommandlockouts

\begin{document}

% paper title
\title{
\MethodAcronym: Open-World Mobile Manipulation via Structured Affordance Grounding
}

% You will get a Paper-ID when submitting a pdf file to the conference system

\def\cameraready{0}  % Comment this out for the arxiv version.

\author{
Kuan Fang$^{*}$,
Yuxin Chen$^{*}$,
Xinghao Zhu$^{*}$,
Farzad Niroui,
Lingfeng Sun,
Jiuguang Wang
\thanks{
$^{*}$Equal contribution. 
This work was conducted at the RAI Institute.
}
\vspace{0.3cm} 
\\ 
\url{https://robot-saga.github.io}
\vspace{-0.5cm} 
}

\maketitle
    
\begin{abstract}
We present SAGA, a versatile and adaptive framework for visuomotor control that can generalize across various environments, task objectives, and user specifications. To efficiently learn such capability, our key idea is to disentangle high-level semantic intent from low-level visuomotor control by explicitly grounding task objectives in the observed environment. Using an affordance-based task representation, we express diverse and complex behaviors in a unified, structured form. By leveraging multimodal foundation models, SAGA grounds the proposed task representation to the robot’s visual observation as 3D affordance heatmaps, highlighting task-relevant entities while abstracting away spurious appearance variations that would hinder generalization. These grounded affordances enable us to effectively train a conditional policy on multi-task demonstration data for whole-body control. In a unified framework, SAGA can solve tasks specified in different forms, including language instructions, selected points, and example demonstrations, enabling both zero-shot execution and few-shot adaptation. We instantiate SAGA on a quadrupedal manipulator and conduct extensive experiments across eleven real-world tasks. SAGA consistently outperforms end-to-end and modular baselines by substantial margins. Together, these results demonstrate that structured affordance grounding offers a scalable and effective pathway toward generalist mobile manipulation.
\end{abstract}

\IEEEpeerreviewmaketitle

\input{1-introduction}
\input{2-related-work}
\input{3-problem-formulation}
\input{4-method}
\input{5-experiments}
\input{6-conclusion}
\input{7-acknowledgement}

% \newpage

{
\small
\bibliographystyle{IEEEtran}
\bibliography{references_lite}
}

\end{document}

%% file: 1-introduction.tex
\section{Introduction}

\begin{figure}[t]
    \centering
    \includegraphics[width=0.5\textwidth]{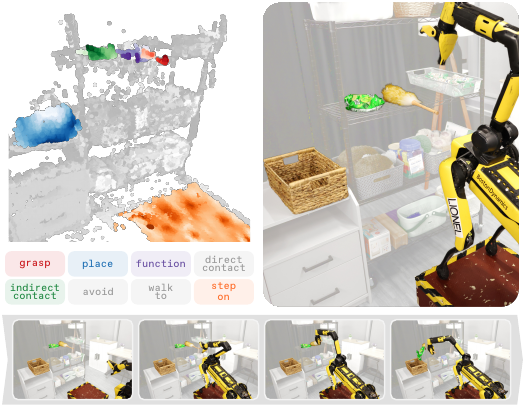} 
    \caption{
    \textbf{\MethodAcronym} expresses diverse, complex mobile-manipulation behaviors using an affordance-based task representation. By explicitly grounding task objectives as 3D heatmaps in the observed environment, our approach disentangles semantic intents from visuomotor control, enabling generalization across environments, task objectives, and user specifications.}
    \label{fig:intro}
    \vspace{-18pt}
\end{figure}

Generalist robots need to seamlessly integrate semantic and geometric understanding to solve diverse and complex tasks in unstructured environments. 
In mobile manipulation~\cite{khatib1999mobile, thakar2023survey} in particular, performing a single task may require concurrent or sequential interactions with multiple objects of different affordances.
An example is shown in \Cref{fig:intro}, where a robot is tasked with retrieving snack bags from a shelf using a duster as a tool. 
During execution, the robot must select actions to achieve the task objectives while accounting for the geometry and configuration of surrounding objects. 
The difficulty is further compounded by the wide range of ways in which users specify task objectives, ranging from natural language to example trajectories, and the variations in how these specifications are expressed.
Achieving such broad generalization across environments, objectives, and specifications remains a central challenge for modern robotic systems.

Recent advances in multimodal foundation models~\cite{achiam2023gpt, team2023gemini, radford2021clip} have created unprecedented opportunities for open-world robotics.
These models can perform strong visual recognition and semantic reasoning over an open set of concepts, yet still lack nuanced physical understanding required for control.
To close the perception-action loop, end-to-end robot foundation models have been trained to directly fuse visual observations with high-level user specifications~\cite{brohan2023rt, black2024pi_0}. 
However, such models must implicitly learn to parse abstract concepts (\eg, ``fluffy duster,'' ``maroon stair''), ground them to raw sensory input, and generate control signals within a black-box model. 
As a result, their generalization capabilities depend on prohibitively large datasets that attempt to span the combinatorial diversity of real-world scenarios, often leading to sharp performance degradation when deployed outside their training distributions.
Alternatively, modular frameworks adopt a more structured design, leveraging pre-trained multimodal foundation models for high-level reasoning, while resorting to hand-engineered modules for low-level execution~\cite{huang2023voxposer, shen2023F3RM}. 
Although more data-efficient, most of these frameworks are less robust in unstructured environments and often constrained to narrowly defined behaviors, such as grasping, limiting their application to sophisticated domains like mobile manipulation. 
Together, these limitations highlight the need for a new paradigm that can retain the open-world reasoning capabilities of foundation models while enabling robust, data-efficient visuomotor control in complex mobile manipulation settings.

% \lingfeng{Following previous comment, this paragraph emphasizes motivation for a representation that is more efficient than end-to-end and more flexible than modularized. Do we need another paragraph to show motivation from a better representation that allows generalization from different user interfaces? like end-to-end models can only take new language instruction, but can't easily use a few new task demos, or user clicks.}

In this work, we present \MethodName (\MethodAcronym), a versatile and adaptable framework for open-world mobile manipulation. To enable broad generalization, our key insight is to disentangle high-level semantic intent from low-level visuomotor control by explicitly grounding task objectives in visual observations. As illustrated in \cref{fig:intro}, we express each task using a set of affordance–entity pairs that specify what to interact with and how the interaction should be performed. Leveraging multimodal foundation models~\cite{achiam2023gpt, radford2021clip}, \MethodAcronym grounds this structured task representation into 3D space as affordance heatmaps. These grounded representations focus the downstream policy on desired behaviors in the context of relevant objects while abstracting away spurious semantic or visual variations that impede generalization, enabling data-efficient learning across a wide range of mobile manipulation tasks.
Using the proposed task representation as a unified interface, \MethodAcronym supports visuomotor control specified in various forms, including instructions, points, and demonstrations, enabling both zero-shot execution and few-shot adaptation for diverse and complex tasks.

We instantiate \MethodAcronym on a quadrupedal mobile manipulation platform operating in cluttered real-world environments. Trained on multi-task demonstration trajectories, \MethodAcronym efficiently learns to solve diverse and complex mobile manipulation tasks without requiring extensive robot data. Across eleven real-world tasks evaluated in zero-shot and few-shot settings, \MethodAcronym exhibits strong generalization and consistently outperforms competitive baselines by substantial margins.

In summary, the key contributions of this work are threefold. 
First, we introduce a structured, affordance-based task representation that unifies diverse task objectives and user specifications. 
Second, we propose a heatmap-conditioned visuomotor control algorithm that grounds task objectives in the 3D space, enabling data-efficient and robust policy learning on multi-task robot data. 
Finally, we instantiate and evaluate this framework on a quadrupedal manipulator, demonstrating strong generalization in unseen real-world tasks. 
Together, these contributions of \MethodAcronym advance the vision of open-world robotic control for mobile manipulation.

%% file: 2-related-work.tex
\vspace{-5pt}
\section{Related Work}

\noindent\textbf{Open-world robotic control.}
Recent advances in foundation models have expanded the frontier of open-world problem solving~\cite{achiam2023gpt, team2023gemini}, motivating growing efforts to leverage large language models (LLMs) and vision-language models (VLMs) for robotic control~\cite{kawaharazuka2024real, yenamandra2023homerobot}. While these models excel at semantic reasoning and compositional planning, they lack fine-grained spatial and physical understanding, making them unreliable for direct visuomotor control.
To mitigate this gap, recent work has explored training generalist robot policies and vision-language-action (VLA) models on large-scale robot datasets~\cite{brohan2022rt, brohan2023rt, kim2024openvla, black2024pi_0, bjorck2025gr00t}. Despite encouraging results, such end-to-end methods demand massive amounts of robot data and remain difficult to scale.
An alternative direction employs prompt engineering and in-context learning to repurpose pre-trained VLMs as high-level planners~\cite{ahn2022can, liang2023code, huang2023voxposer, fangandliu2024moka, liu2025dynamem}, to generate textual subtasks that can be executed by hand-designed controllers. However, these pipelines require extensive manual tuning, and the fixed primitives are highly sensitive to prediction errors from the high-level model.
Our work instead introduces a spatially grounded, affordance-based task representation as a unified interface between high-level semantic reasoning and low-level visuomotor control. Using affordance heatmaps as task conditioning, our approach efficiently learns generalizable policies in the real world.

\noindent\textbf{Task representations for robotics.}
A central challenge in learning generalist robot policies lies in designing task representations that are both expressive and flexible. Traditional goal-conditioned policies specify tasks through goal states or observations, which in principle can describe arbitrary objectives given an appropriate goal space~\cite{kaelbling1993learning, liu2022goal}. However, they rely on precise goal specification, which is often difficult for users to provide in the real world. Alternatively, language-based task specifications provide a natural and general interface for humans~\cite{lynch2020language, walke2023bridgedata}, but their symbolic nature makes it difficult to ground semantic intent in the robot’s high-dimensional sensory inputs. While several works have sought to train goal- and language-conditioned visuomotor policies end-to-end, their applicability and scalability in open-world settings remain limited.
To overcome these challenges, recent studies have explored spatially grounded task representations such as heatmaps~\cite{shridhar2022cliport, kerr2023lerf, shen2023F3RM, wang2023d}, masks~\cite{fang2018mtda, yin2025codediffuser}, keypoints~\cite{fangandliu2024moka, yuan2024robopoint, huang2024rekep}, flows~\cite{xu2024im2flow2act, yuan2024generalflow, Gu2023RTTrajectoryRT}. These representations link high-level reasoning and low-level control, providing clear spatial and geometric cues as inputs to the policy. Building on this insight, we introduce a structured, affordance-based representation that encodes task objectives and ground them as continuous affordance heatmaps in the 3D space. In contrast, our formulation covers a broader range of behaviors for mobile manipulation by composing a set of affordances. Moreover, it supports both zero-shot execution and few-shot adaptation and achieves superior performances in diverse and complex tasks.

\noindent\textbf{Mobile manipulation.}
A substantial body of research has explored the integration of manipulation and mobility across diverse embodiments~\cite{khatib1999mobile, sentis2006whole, asfour2006armar, thakar2023survey, gu2025humanoid, 2025relic}.
Early work primarily relied on model-based control~\cite{khatib1999mobile, sentis2006whole, berenson2008optimization, bohren2011towards, ciocarlie2012mobile}, in which robot kinematics and dynamics were carefully modeled for specific platforms and tasks. 
% While these approaches offered elegant control formulations, they often lacked robustness and scalability in unstructured environments.
\cite{garrett2021integrated, ding2023task} integrate symbolic task planning with motion optimization for compositionality and interpretability. However, they typically rely on handcrafted low-level planners, symbolic preconditions, accurate geometric models, limiting scalability and adaptability in open-world settings.
In contrast, our approach focuses on learning robust visuomotor control to handle unstructured environments and achieves compositional reasoning through structured affordance grounding.
Recent studies have adopted data-driven methods through reinforcement and imitation learning~\cite{li2020hrl4in, wu2023tidybot, zimmermann2021go, sun2022fully, fu2023deep, fu2024mobile, kareer2025egomimic}. Although these approaches have shown promising results, they typically demand extensive training data, limiting their applicability to a simple behaviors such as object search and rearrangement.
Our work advances versatile and generalist control by introducing a unified interface that enables a mobile manipulator to perform diverse behaviors.

\begin{figure*}[t]
    \centering
    \includegraphics[width=\textwidth]{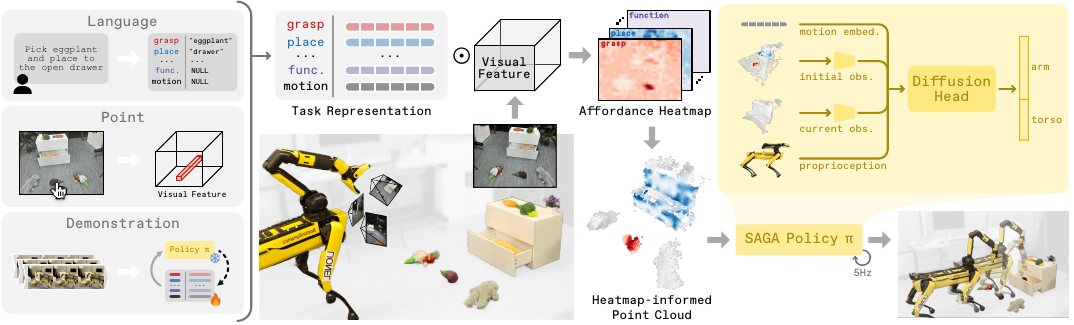} 
    \vspace{-10pt}
    \caption{
    \textbf{Overview of \MethodAcronym.} Given the user specification, provided via language instructions, selected points, or example demonstrations, \MethodAcronym computes the task representation consists of a set of affordance–entity pairs in latent space. Based on their similarity with visual embeddings, the task representation is grounded to visual observations to form 3D affordance heatmaps, which guides a conditional policy to predict the actions. Using the proposed task representation as a unified interface, \MethodAcronym enables the robot to achieve both zero-shot execution and few-shot adaptation for open-world mobile manipulation.
    }
    \vspace{-15pt}
    \label{fig:method}
\end{figure*}

%% file: 3-problem-formulation.tex
\section{Problem Formulation}
\label{sec:problem_formulation}

We consider the problem of mobile manipulation in unstructured environments, where a robot is commanded to interact with objects based on the high-level user specification. The robot receives the observation composed of onboard RGB-D views and proprioceptive states, and produces the whole-body action that jointly controls torso and arm motion.

In this work, we allow the user specification to be given in one of three common forms:  
\emph{Language} instructions offer the most general and expressive way to specify task objectives.  
\emph{Point} inputs allow the user to designate relevant entities by selecting regions in the robot’s visual observations, providing direct and precise guidance.  
\emph{Demonstration} consists of one or a few example trajectories that illustrate the desired behavior when a text or spatial description is unavailable.

Our objective is to learn a policy $\pi$ that computes the action $a$ based on the observation $o$ and the user specification $u$. Following \cite{walke2023bridgedata, black2024pi_0}, we train this policy through imitation learning on real-world, multi-task demonstrations with annotated language instructions. At test time, the robot executes or adapts to novel tasks involving previously unseen environments, task objectives, and user specifications.

%% file: 4-method.tex
\section{Method}

We present \MethodName (\MethodAcronym), a framework for versatile and  adaptive mobile manipulation by grounding task objectives explicitly in the 3D space. Achieving this requires addressing several key challenges. First, how to represent task objectives in a structured form while covering diverse behaviors. Second, how to ground this representation into the robot’s observations to decouple high-level semantics from low-level visuomotor control. Third, how to robustly generate actions conditioned on the grounded representation. Fourth, how to support tasks specified through diverse user inputs. Finally, how to deploy this framework on real mobile manipulation systems.

% --------------------------------
\subsection{Affordance-Entity Pairs as Task Representation}
\label{sec:task_representation}

To support broad generalization across task objectives, we propose a structured task representation that effectively expresses diverse and complex physical interactions in a unified manner. We represent the objectives of each task using a set of affordance–entity pairs that specify \textit{what} the robot should interact with and \textit{how} the interaction should be conducted. For example, the sweeping task in \Cref{fig:intro} can be expressed as a dictionary \{grasp: ``duster handle'', function: ``duster head'', indirect\_contact: ``snack'', place: ``woven basket'', step\_on: ``maroon stair''\}. In contrast to prior work~\cite{shen2023F3RM, liu2024visual} that focuses only on narrowly scoped skills, \MethodAcronym spans a set of affordance types beyond grasping, enabling flexible composition of multiple objectives within the same formulation. 

Formally, the set of affordance-entity pairs are encoded as $\{(w_k, z_k)\}_{k=1}^K$, where $w_k$ is one of $K$ affordance types and $z_k \in \mathbb{R}^M$ is a $M$-dimensional entity embedding. Each embedding characterizes the semantic properties of the entity for identifying their location and spatial extent in the visual observation of the environment. These embeddings can be obtained from language or visual descriptions extracted from the user specification $u$ using a pretrained multimodal encoder $\psi(\cdot)$~\cite{radford2021clip}, which will be detailed \Cref{sec:versatile_interface}. Embeddings are set to zero if the corresponding affordance is irrelevant for the task, ensuring a fixed-dimensional task representation.

While such affordance–entity pairs typically capture most essential task objectives, additional information might need to be specified for certain behaviors. For instance, a sweeping motion might require specifying not only the target entities but also a motion direction ``from right to left''. Thus, we augment $c$ with a motion embedding $z_{\text{motion}}$ computed using the motion information extracted from $u$. Since the affordance–entity pairs already encode the primary semantics, this motion embedding remains compact yet ensuring the expressiveness of $c$. As shown in \Cref{fig:method}, the complete task representation becomes:
\begin{equation}
    c = [z_1, \ldots, z_K, \, z_{\text{motion}}].
    \label{eqn:task_representation}
\end{equation}
For long-horizon tasks with multiple stages, we follow~\cite{huang2023voxposer, fangandliu2024moka} to decompose the task into a sequence of subtasks $[c_1,c_2, \dots]$, where each element is represented same as in \Cref{eqn:task_representation}. 

Now we have a unified, entity-centric representation that covers broad task objectives. Next, we explain how this representation enables generalizable visuomotor control through spatial grounding, while deferring how $c$ is computed from different user specifications to \Cref{sec:versatile_interface}.

% --------------------------------
\subsection{Structured Affordance Grounding}
\label{sec:heatmaps}

Robust visuomotor control requires grounding the task representation to the robot's observation. Although expressive, the entity embeddings in this representation often contain detailed semantic or visual information irrelevant to physical interaction, which can hinder generalization if supplied directly to the policy. For instance, variations in texture or phrasing (\eg, “yellow duster” \textit{vs.} “fluffy cleaning tool”) should not affect the intended motion, yet their latent embeddings can differ substantially.
Instead of directly predicting actions based on the entity embeddings, we convert each of them into an affordance heatmap that marks the spatial information of the target entity for each affordance type. This grounding preserves the fine-grained structure of the task objectives while abstracting away nonessential semantics.

Inspired by \cite{shen2023F3RM, liu2024visual}, we compute the heatmap by encoding the visual observation into the same latent space with the entity embeddings and measuring their similarity. In contrast to focusing on grasping only, we compute a multi-channel heatmaps for a compositional set of affordance types.
Given an RGB-D image in the observation $o$, we extract visual embedding $\psi(o) \in \mathbb{R}^{W \times H \times M}$ using the same pretrained multimodal encoder for producing the entity embeddings. For each affordance type $w_k$ with entity embedding $z_k$, we compute an affordance heatmap by cosine similarity:
\begin{equation}
    h_{k}^{i} = \frac{\psi(o)^i \cdot z_k}{\|\psi(o)^i\| \, \|z_k\|},
\label{eq:heatmap}
\end{equation}
where $h_k^i$ reflects how strongly pixel $i$ corresponds to the affordance associated with $w_k$. As shown in \Cref{fig:method}, stacking across $K$ affordance types yields a $W \times H \times K$ tensor as the heatmap, representing the grounded task semantics on the 2D visual observation of the environment. 

To tightly align the heatmap with the geometry of the environment, we lift the heatmap into 3D along with the point cloud $x$ computed from the depth channel from $o$. Each 3D point is thus associated with a $K$-dimensional affordance feature, forming a heatmap-informed point cloud $[x, h]$. This grounds task objectives to the environment in a structured manner for the downstream visuomotor control.

% --------------------------------
\subsection{Heatmap-Conditioned Visuomotor Control}
\label{sec:policy_architecture}

Unlike end-to-end policies that directly combine raw RGB images with high-level user specifications~\cite{brohan2022rt, black2024pi_0}, the \MethodAcronym policy operates on the heatmap-informed point cloud. This design enables the policy to focus on the spatial and geometric information needed for physical interactions, leading to efficient generalization across diverse scenarios.

A major challenge for the policy is maintaining consistent grounding as the environment evolves. A straightforward design would recompute affordance heatmaps at every timestep $t$ from the latest observation $o_t$. However, this would require repeatedly running the heavy multimodal encoder and can often become brittle once objects self-occlude during execution. Instead, we compute the affordance heatmaps $h = h_0$ once from the initial observation $o_0$, which typically provides a clean and complete view of the scene. As the robot and objects move, the policy learns to implicitly align $o_0$ and $o_t$ through their shared point cloud structure, maintaining spatio-temporal correspondence without regenerating heatmaps. 

Formally, the policy is denoted as $\pi(a_{t:t+T-1} \mid c, o_0, o_t)$, where a $T$-step action chunk~\cite{chi2023diffusion} is predicted at each timestep to ensure temporal consistency and mitigate compounding error. We instantiate $\pi$ as a conditional diffusion policy using a two-stream PointNet encoder~\cite{qi2017pointnet}. One stream embeds the heatmap-informed point cloud $[x_0, h]$, capturing globally grounded task semantics. The other stream embeds the current point cloud $x_t$, capturing the local geometry required for real-time interaction. Their features are fused with the motion embedding contained in $c$ and the proprioceptive state  to produce a latent representation encoding both what needs to be achieved and how the scene is changing. A diffusion head is applied at the end to predict the $T$-step action chunk $a_{t:t+T-1}$ to perform closed-loop control.

We train the policy through conditional imitation learning, which leverages multi-task demonstration data to efficiently align actions with diverse specifications. During training, an annotated multi-task dataset is provided as $\mathcal{D} = \{ \tau^j \}_{j=1}^{|\mathcal{D}|}$.
Each trajectory $\tau^j$ in the dataset consists of the sequence of observations $o_t^j$ and actions $a_t^j$ as well as the task representation $c^j$, which is computed from annotated text descriptions. 
To reduce over-sensitivity to perception and specification variations, heatmap augmentation is applied during training by randomly rescaling and sharpening each channel of the computed heatmap.
This encourages the policy to focus on spatial and semantic structures rather than exact heatmap magnitudes, mitigating brittleness to shifts in task phrasing, encoder error, or sensing noise.

% --------------------------------
\subsection{Versatile Interfacing to User Specifications}
\label{sec:versatile_interface}

A major advantage of \MethodAcronym is that its structured task representation serves as a unified, modality-agnostic interface for specifying user intent. As shown in \Cref{fig:method}, we employ the trained \MethodAcronym policy for the three common modalities of user specifications described in \Cref{sec:problem_formulation}, spanning both zero-shot execution (language, point) and few-shot adaptation (demonstration), to demonstrate its versatility.

\noindent
\textbf{Language.}
Following~\cite{fangandliu2024moka}, we employ a VLM~\cite{achiam2023gpt} to decompose the instruction into a sequence of subtasks, and extract the text description for the motion and target entities. Using the multimodal encoder~\cite{radford2021clip}, these texts are converted into entity embeddings and the motion embedding. By outsourcing high-level semantic reasoning and visual recognition to pretrained foundation models, \MethodAcronym can perform diverse and complex physical interactions for an open set of objects and task goals.

\noindent
\textbf{Point.}
Given the selected pixel location $p_k$ for each affordance type $w_k$, the corresponding visual embedding $\psi(o_0)^{p_k}$ can naturally serve as the entity embedding $z_k$ for the specified affordance $w_k$. Note that the points need not be precisely specified on the exact position where the robot should grasp or contact the object, as different parts of the same object usually share similar embeddings for a well trained encoder. To further improve the robustness, we compute $z_k$ as the average over a local $3\times 3$ window centered around $p_k$. This enables intuitive and convenient user interface without requiring language parsing or policy fine-tuning.

\noindent
\textbf{Demonstration.}
Instead of directly fine-tuning the policy, which would typically require hundreds of demonstrations, we freeze the pre-trained policy and only optimize the task representation over the few-shot demonstration $\mathcal{D}'$: 
\begin{equation} 
    c^{*} = \arg \min_{c} \sum_{\tau \sim \mathcal{D}'} \! -\log \pi(a_{t:t+T-1} \mid c, o_0, o_t). 
\end{equation}
Because the mapping $c \rightarrow h \rightarrow a$ is fully differentiable, the optimization can be effectively conducted via backpropagation to embeddings in $c$, analogous to soft prompt tuning~\cite{jia2022visual}. This novel paradigm, which we refer to as \emph{heatmap tuning}, enables few-shot adaptation without ground truth instructions while retaining the capabilities of the pre-trained policy. 

% --------------------------------
\subsection{Mobile Manipulation System Summary}
\label{sec:system_summary}

\MethodAcronym is instantiated on a quadrupedal manipulator as illustrated in \Cref{fig:method}. We summarize the key components of this instantiation in details below.

\noindent
\textbf{Robot platform.}
We deploy \MethodAcronym on a Spot robot equipped with a 6-DoF arm and a parallel-jaw gripper~\cite{bostondynamics_spot}. One wrist-mounted and two forward-facing cameras provide multi-view RGB-D observations with known extrinsic and intrinsic parameters. The observation $o_t$ also includes a 19-dimensional proprioceptive state encoding torso pose, end-effector pose, and finger position. Following~\cite{chi2023diffusion}, each $SE(3)$ pose is represented as a 9-dimensional vector. The 21-dimensional action specifies 9-dimensional target poses for the torso and the end-effector, together with binary flags as defined in~\cite{brohan2022rt}.

\noindent
\textbf{Affordance types.}
Based on \cite{fangandliu2024moka}, we consider eight affordance types that span core mobile manipulation capabilities as described in \Cref{table:affordance_types}. These affordances can be combined either sequentially or concurrently to express complex objectives. For example, a task may require \textit{grasp} a mug and \textit{place} it on a rack while \textit{avoiding} a laptop. Importantly, \MethodAcronym supports both under-specified and over-specified commands thanks to the expressiveness of the learned policy. For instance, a user may specify only \textit{grasp} when the mug is distant, or additionally include \textit{walk to} the table, with the trained policy resolving the ambiguity based on the environment context. 

\noindent
\textbf{Model and training.}
To ensure spatial consistency across time, all observations and actions are expressed in the same frame, centered at the end-effector at $t$. At runtime, RGB-D streams from all cameras are independently converted to point clouds and affordance heatmaps then fused together. The fused cloud is cropped to a $2\,\text{m}$ workspace and uniformly downsampled to $N{=}1024$ points for real-time inference. The policy network and pretraining follow~\cite{ze20243d}, while few-shot adaptation optimizes only the task representation $c$ with an elevated learning rate of $1\times10^{-3}$ to enable fast convergence.

%% file: 5-experiments.tex
\section{Experiments}
We conduct extensive experiments and analyses to evaluate the effectiveness of \MethodAcronym. 
Specifically, we aim to study the following questions:
\textbf{Q1:} Does the proposed task representation effectively capture diverse task objectives and ground them reliably in the environment?
\textbf{Q2:} How well does \MethodAcronym generalize to novel tasks and environments in zero-shot manners compared to state-of-the-art baselines?
\textbf{Q3:} Can the unified task representation support different forms of user specification and enable fast adaptation?

\begin{table}[t]
    \centering
    \caption{
    \textbf{Affordance types.}
    We consider eight affordance types to express task objectives. Used individually or compositionally, they express a wide range of physical interactions required for robust mobile manipulation.
    }
    \rowcolors{2}{gray!10}{white}
    \begin{tabular}{l|p{0.65\columnwidth}}
    \textbf{Affordance} & \textbf{Definition} \\ \toprule
    grasp            & Entity to be grasped by the gripper \\
    place            & Target region for placing the grasped object \\
    function         & Functional part of the grasped object \\
    direct\_contact   & Scene entity directly contacted by the gripper \\
    indirect\_contact & Scene entity contacted by the function entity \\
    avoid            & Entity the robot must not contact or traverse \\
    walk\_to          & Entity to approach and bring within gripper reach \\
    step\_on          & Entity for the robot to set foot on 
    \end{tabular}
    \vspace{-5pt}
    \label{table:affordance_types}
\end{table}

\begin{figure}[t]
    \centering
    \includegraphics[width=\linewidth]{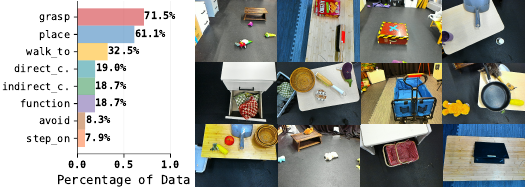}
    \caption{
    \textbf{Training data.} The \MethodAcronym policy is trained on 2,410 demonstration trajectories collected across diverse scenes encompassing various combinations of eight affordance types. Example scenes are shown, along with the marginal distribution of demonstrations containing each affordance type.
    }
    \vspace{-18pt}
    \label{fig:data}
\end{figure}

\begin{figure*}[t]
    \centering
    \includegraphics[width=\textwidth]{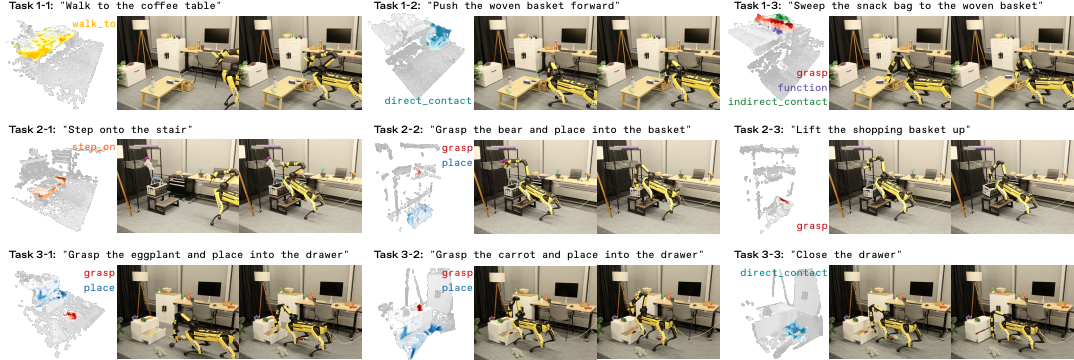} 
    \caption{
    \textbf{Task execution.} We deploy and evaluate \MethodAcronym for mobile manipulation in cluttered environments of novel objects and configurations. 
    Results are organized in a $3\times3$ grid, where each row shows the three tasks for the same long-horizon scenario. Each entry shows the affordance heatmap (left) and 2 frames of task execution (right).
    \MethodAcronym successfully grounds the task objectives and achieves high success rates in all tasks, exhibiting strong generalization and robustness.
    }
    \label{fig:experiment_main_task_execution}
    \vspace{-15pt}
\end{figure*}

% --------------------------------
\subsection{Experimental Setup}
\label{sec:experimental_setup}

We evaluate \MethodAcronym on real-world mobile manipulation tasks to assess its generalizability, robustness, and adaptability.

\noindent
\textbf{Training data.} 
As shown in \cref{fig:data}, we collect 2{,}410 demonstration trajectories via teleoperation, covering a diverse set of behaviors with annotated affordances, including pulling a cart~(grasp, walk\_to), poking with a shovel~(grasp, function, indirect\_contact), closing a laptop~(direct\_contact), \etc. Object instances, spatial configurations, and scene layouts are randomized across environments to promote broad generalization. Each trajectory contains up to 600 steps, resulting in approximately 1.3M state–action pairs in total. Notably, this dataset is two orders of magnitude smaller than those used by prior generalist robot policies~\cite{black2024pi_0}, underscoring the substantially higher efficiency of our approach.

\noindent
\textbf{Testing tasks.} We construct testing environments resembling household, office, and retail spaces, each containing unseen furniture and object instances. Across these environments, we define three long-horizon tasks, each decomposed into three sequential sub-tasks, yielding nine evaluation tasks as shown in \Cref{table:testing_tasks}. Each task is indexed as $i$–$j$, where $i$ denotes the scenario and $j$ the sub-task. To assess heatmap-tuning, we additionally design two tasks requiring object-level and task-level adaptations as described in \cref{sec:experiments_few_shot}.

\begin{table}[t]
    \centering
    \caption{
    \textbf{Testing tasks.} 
    We evaluate 9 sub-tasks organized into 3 long-horizon mobile manipulation scenarios. MPNP, TPNP, and HPNP denote Mobile Pick-and-Place, Table-top Pick-and-Place, and Horizontal Pick-and-Place, respectively.
    }
    \rowcolors{2}{white}{gray!10}
    \begin{tabular}{p{0.20\columnwidth}|p{0.70\columnwidth}}
    \textbf{Task} & \textbf{Description} \\ \toprule
    1-1 \textit{Walk To} & Walk to a small table \\ 
    1-2 \textit{Push} & Push a basket beside the table using the gripper \\ 
    1-3 \textit{Sweep} & Sweep a snack into the basket with a brush \\ \midrule
    2-1 \textit{Step On} & Approach the shelf and step onto the stair \\ 
    2-2 \textit{HPNP} & Pick horizontally from shelf and place in basket \\ 
    2-3 \textit{Lift} & Lift up the shopping basket from the ground \\ \midrule
    3-1 \textit{MPNP} & Walk to an object and place it into the open drawer \\ 
    3-2 \textit{TPNP} & Pick up an object and put it into the open drawer \\ 
    3-3 \textit{Close} & Close the open drawer with the gripper \\ 
    \end{tabular}
    \vspace{-15pt}
    \label{table:testing_tasks}
\end{table}

\noindent
\textbf{Baselines.} We compare \MethodAcronym with four baseline methods. \textit{DP3}~\cite{ze20243d} is a diffusion policy originally designed for task-specific training. We convert it to a multi-task policy using the embedding of the language instruction computed by \cite{radford2021clip} as the task representation. \textit{CodeDiffuser}~\cite{yin2025codediffuser} explicitly extracts entity descriptions from the original language instruction using a VLM and uses binary masks to exclude distractor objects from the input point clouds. 
\textit{$\pi_0$}~\cite{black2024pi_0} trains a VLA model end-to-end based on the input RGB images and langauge instructions. We further include depth images to its inputs for fair comparison and fine-tune the model on our collected dataset. 
For few-shot adaptation, we additionally compare with \textit{SKIL}~\cite{wang2025skil}, which adapts to new tasks using learned keypoint representations. 
CodeDiffuser is excluded from few-shot adaptation and SKIL from zero-shot execution, as their formulations do not directly support those settings. 
All methods are trained and evaluated following the same protocols for fair comparison.

% --------------------------------
\begin{figure*}[t]
    \centering
    \includegraphics[width=\linewidth]{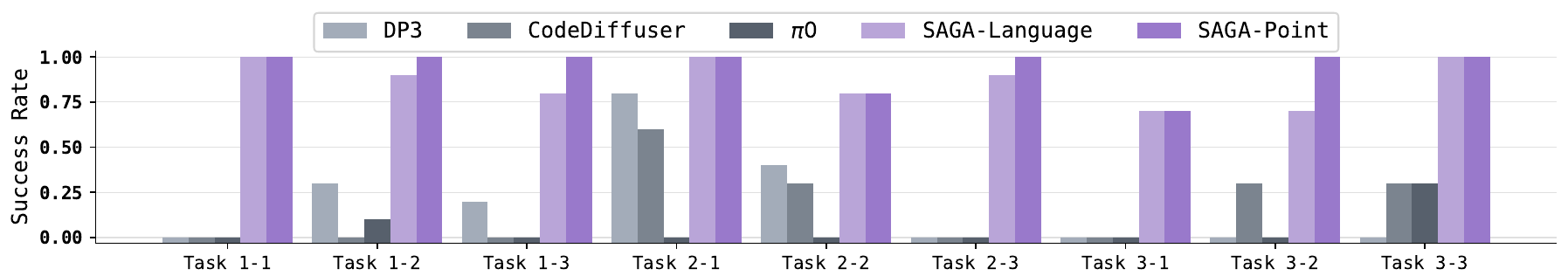}
    \vspace{-15pt}
    \caption{
    \textbf{Zero-shot execution performance.} 
    We evaluate \MethodAcronym\ on nine unseen mobile manipulation tasks and report average success over 10 trials. Both variants, \MethodAcronym-Language and \MethodAcronym-Point, consistently outperform baselines, achieving superior success across a wide range of tasks.
    These results demonstrate the robust zero-shot execution of our approach.
    }
    \vspace{-10pt}
    \label{fig:zero_shot_performance}
\end{figure*}

% --------------------------------
\subsection{Zero-Shot Execution}
\label{sec:zero_shot}

We first evaluate all methods conditioned on natural language instructions and additionally evaluate a variant of our method using the point specifications. We denote our model variants as \textit{\MethodAcronym-Language} and \textit{\MethodAcronym-Point}. 
The average success rates across 10 trials are reported in \Cref{fig:zero_shot_performance}. 
Both \MethodAcronym variants achieve high success rates across all tasks.
Even in tasks composing multiple affordance types (\eg, sweeping with previously unseen tools), \MethodAcronym maintains strong performance. 
Moreover, given the same affordance types (\eg, grasp and place), the trained policy can behave differently in accordance with different environment contexts, performing top-down, horizontal, or mobile grasping respectively.
Between the two variants, \MethodAcronym-Point achieves modestly higher success rates, as a selected point directly identifies the region of interest and resolve semantic ambiguity, which is particularly helpful when multiple objects share similar semantics. 

In contrast, baselines lacking structured task representations exhibit systematic failure patterns. 
DP3 and CodeDiffuser frequently mislocalize the target objects, leading to unstable grasps and incorrect contacts. 
While CodeDiffuser uses a VLM to segment target objects, its binary mask representation combines all target objects together without distinguishing how each object should be interacted with, resulting in ambiguous intents. 
Despite extensive pretraining, the end-to-end trained $\pi_0$ does not effectively adapt to the quadrupedal manipulator, which is unseen in its pre-training dataset, due to its massive model size and the relatively small fine-tuning data (less than 0.2\% of the original dataset).
Consequently, its output actions exhibit mode collapse, only occasionally succeeding on less complex tasks (\eg, \textit{Push}, \textit{Close}).
These results highlight the advantages of structured affordance grounding for efficiently learning robust mobile manipulation in open-world settings.

% --------------------------------
\subsection{Few-Shot Adaptation}
\label{sec:experiments_few_shot}

We next evaluate whether \MethodAcronym can adapt to novel tasks using only 10 demonstrations via the heatmap-tuning procedure introduced in \cref{sec:versatile_interface}. We consider two representative settings: 
(i) \textbf{object-level adaptation} (\textit{sort vegetable}), where an in-distribution affordance set (\eg, \textit{grasp}, \textit{place}) is applied to previously unseen objects, 
and (ii) \textbf{task-level adaptation} (\textit{clean-avoid}), where the robot is asked to solve the task specified by a novel combination of affordance types (\textit{grasp}, \textit{avoid}, and \textit{indirect\_contact}) unseen during policy training. 

\begin{figure}[t]
    \centering
    \includegraphics[width=\linewidth]{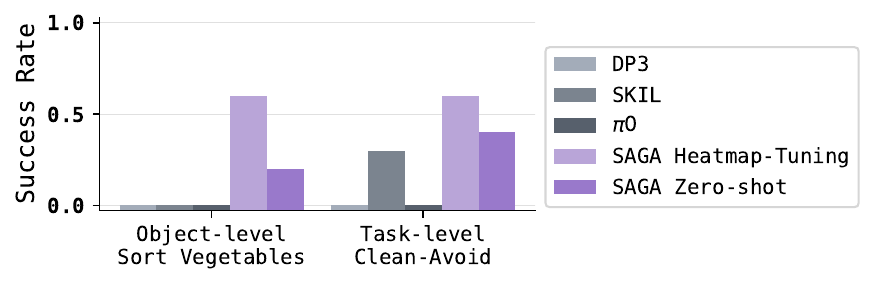}
    \vspace{-20pt}
    \caption{
    \textbf{Few-shot adaptation performance.}
    We evaluate \MethodAcronym for object-level and task-level adaptation using 10 demonstrations. Through heatmap-tuning, \MethodAcronym consistently outperforms baselines and the zero-shot model variant.
    }
    \vspace{-15pt}
    \label{fig:few_shot_performance}
\end{figure}

\begin{figure}[t!]
    \centering
    \includegraphics[width=\linewidth]{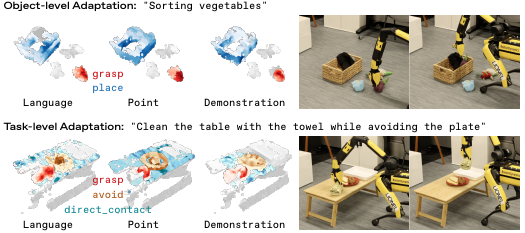}
    \caption{
    \textbf{Qualitative results of few-shot adaptation.} For each task evaluated in \cref{sec:experiments_few_shot}, we visualize the affordance heatmaps computed from different user specification modalities on the left and the task execution on the right.
    }
    \vspace{-15pt}
    \label{fig:few_shot_task_execution}
\end{figure}

We evaluate \MethodAcronym and baselines in few-shot manners without language instructions, as well as a zero-shot variant of \MethodAcronym given ground-truth instructions. As shown in \cref{fig:few_shot_performance}, \MethodAcronym achieves reasonable zero-shot successes and rapidly improves success rates through heatmap tuning, reaching stable and reliable execution with only ten demonstrations. By optimizing the task representation while keeping the visuomotor policy frozen, the adapted task representations highlight the relevant affordance regions on the point cloud, leading to affordance heatmaps of the quality comparable to ground truth heatmaps computed from instructions and points, as shown in \Cref{fig:few_shot_task_execution}.
In contrast, all baselines perform poorly in zero-shot and struggle to adapt effectively. DP3 and SKIL can sometimes approach and grasp objects but exhibit unstable trajectories, leading to inconsistent performance and frequent task failure.

%% file: 6-conclusion.tex
\section{Conclusion and Discussion}
\label{sec:conclusion}

We presented \MethodAcronym, a unified framework for open-world mobile manipulation that explicitly grounds task objectives in 3D geometry. By representing tasks as affordance–entity pairs and mapping them into affordance heatmaps, \MethodAcronym decouples high-level semantic reasoning from low-level visuomotor control. This structured grounding enables a single conditional policy to robustly perform diverse tasks across varying environments and objectives. Moreover, the proposed task representation serves as a modality-agnostic interface, allowing the trained policy to be conditioned from language instructions, mouse clicks, or example demonstrations. Extensive real-world evaluations on a quadrupedal manipulator demonstrate strong generalization, robust task execution, and rapid adaptation, significantly outperforming prior end-to-end and modular baselines. These results highlight the promise of spatially grounded task representations for scalable and generalizable robot learning in the real world.

Despite these advances, several limitations suggest promising directions for future work. 
First, the current affordance vocabulary, while expressive for a wide range of tasks, remains tailored to single-arm mobile manipulation. 
Extending to bimanual, dexterous, or humanoid systems will likely require designing or learning affordance types that capture richer interaction semantics. 
Second, while prioritizing spatially grounded affordances and object geometry significantly improves robustness and generalization, complex tasks involving deformable objects or nuanced material properties may benefit from incorporating compact visual cues to complement the affordance-informed point cloud within a unified representation. 
Finally, advances in multimodal encoders and correspondence estimation may enable more reliable online updating of affordance grounding during execution, further improving performance in dynamic and partially observed environments.

%% file: 7-acknowledgement.tex
\section*{Acknowledgment}
\label{sec:acknowledgement}

We would like to thank Jessica Hodgins, Tao Pang, Dawn Wendell, and Dogan Yirmibesoglu for providing feedback on early drafts of the manuscript. 
We also want to thank Simon Le Cleac'h, Andy Park, and Zhaoming Xie for their support on the robot infrastructure and experiments. 